\documentclass[10pt, conference, letterpaper]{IEEEtran}
\usepackage[ruled,linesnumbered,vlined]{algorithm2e}
\usepackage{verbatim}
\usepackage{amsfonts}
\usepackage{epsfig}
\usepackage{algorithmic}
\usepackage{multicol}
\usepackage{multirow}
\usepackage{amsmath}
\usepackage{amssymb}
\usepackage{subfig}
\usepackage{textcomp}
\usepackage{tikz}
\usepackage{pgf}
\usepackage{float}
\usepackage{graphicx}
\usepackage{comment}
\usepackage{mathtools}

\DeclarePairedDelimiter\floor{\lfloor}{\rfloor}

\usepackage{url}      

\usepackage{diagbox}
\usepackage{array}
\newcolumntype{M}[1]{>{\centering\arraybackslash}m{#1}}

\makeatletter
\def\url@leostyle{%
  \@ifundefined{selectfont}{\def\UrlFont{\sf}}{\def\UrlFont{\small\bf\ttfamily}}}
\makeatother
\IEEEoverridecommandlockouts
\begin{document}

\title{Dilated Deep Residual Network for Image Denoising}

\author{Tianyang Wang$^\text{*}$ \thanks{Corresponding author's email: toseattle@siu.edu} \and Mingxuan Sun \and Kaoning Hu

}

\providecommand{\keywords}[1]{\textbf{\textit{Keywords---}} #1}

\maketitle

\begin{abstract}

Variations of deep neural networks such as convolutional neural network (CNN) have been successfully applied to image denoising. The goal is to automatically learn a mapping from a noisy image to a clean image given training data consisting of pairs of noisy and clean images. Most existing CNN models for image denoising have many layers. In such cases, the models involve a large amount of parameters and are computationally expensive to train. In this paper, we develop a dilated residual CNN for Gaussian image denoising. Compared with the recently proposed residual denoiser, our method can achieve comparable performance with less computational cost. 
Specifically, we enlarge receptive field by adopting dilated convolution in residual network, and the dilation factor is set to a certain value. We utilize appropriate zero padding to make the dimension of the output the same as the input. It has been proven that the expansion of receptive field can boost the CNN performance in image classification, and we further demonstrate that it can also lead to competitive performance for denoising problem. 
Moreover, we present a formula to calculate receptive field size when dilated convolution is incorporated. Thus, the change of receptive field can be interpreted mathematically. To validate the efficacy of our approach, we conduct extensive experiments for both gray and color image denoising with specific or randomized noise levels. Both of the quantitative measurements and the visual results of denoising are promising comparing with state-of-the-art baselines.

\end{abstract}


\section{Introduction}
\label{introduction}

Image denoising has been a fundamental yet challenging research topic in computer vision. The goal is to reconstruct a clean image from a noisy observation. Generally, a noisy image is modeled as $y$ = $x$ + $z$, where $x$ is a clean image and $z$ is some type of noise such as the additive Gaussian white noise. Traditional methods such as BM3D \cite{dabov2007image}, LSSC \cite{mairal2009non}, EPLL \cite{zoran2011learning}, and WNNM \cite{gu2014weighted} rely on image prior modeling and highly require domain knowledge. However, finding an effective prior is still difficult.  

Variations of deep neural networks such as convolutional neural network (CNN) have been successfully applied to image denoising \cite{jain2009natural,burger2012image}. The goal is to automatically learn a mapping from a noisy image to a clean image given training pairs. Besides Jain's work \cite{jain2009natural} and MLP \cite{burger2012image}, successful methods in this category also include  CSF\cite{schmidt2014shrinkage}, DGCRF \cite{vemulapalli2016deep}, NLNet \cite{lefkimmiatis2016non}, TNRD \cite{chen2017trainable}, and DP \cite{zhang2017learning}. A deep residual net has been further proposed in~\cite{zhang2017beyond} to learn a residual noisy mapping instead of a clean one and the results outperform most state-of-the-art methods.

Most of the existing CNNs for image denoising have many layers, e.g. $17$ and $20$ in~\cite{zhang2017beyond}. In such cases, the models involve a large amount of parameters and are computationally expensive to train. It has been shown that enlarging receptive field is effective in improving CNN performance. There are a number of ways to expand receptive field. One option is simply stacking more convolutional layers and increasing the depth. However, it inevitably brings more parameters and increases the computational burden. Another option is to adopt pooling operations. However, pooling alone cannot be directly applied to image denoising, since the output and input should have the same dimension. Adding up-convolution \cite{long2015fully} after pooling operation can make the dimension of the output the same as the input, however the performance of denoising may get worse. It is therefore important to design an efficient architecture of fewer layers without sacrificing the performance.

In this paper, we propose a deep residual network with dilated convolution for image denoising. Specifically, our model aims to learn residual noisy mapping inspired by Zhang\textquotesingle s work \cite{zhang2017beyond}. Once the mapping is learned, denoising can be performed by subtracting the noisy mapping from the original noisy input. More importantly, we incorporate dilated convolution to effectively enlarge receptive filed of each convolutional layer. As shown in Figure \ref{figure1}, the fundamental layer pattern of our model is \textquoteleft DilatedConv-BN-ReLU\textquoteright, where \textquoteleft DilatedConv\textquoteright ~refers to dilated convolution with a certain factor, and BN and ReLU refer to batch normalization \cite{ioffe2015batch} and rectified linear unit \cite{maas2013rectifier}, respectively. BN is mainly used for training convergence, and ReLU is to add non-linearity to improve network ability of extracting discriminative features. 

With the aid of dilated operations, our model can achieve comparable denoising capacity without increasing network depth, which is more computational efficient. Specifically, we adopt 2-dilation with smaller filter size, which can expand receptive field more aggressively. We demonstrate the improvement of receptive field expansion and compare our model with the state-of-the-art method (DnCNN) \cite{zhang2017beyond} in section \ref{DenoisingResNet}. Extensive experiments validate that the increment of receptive field size in CNN is capable of capturing important image clues and thus enhances the denoising performance. Furthermore, to better clarify how dilated convolutions impact receptive field, we revisit the convolution arithmetic and propose a formula to calculate receptive field size when dilated convolution is involved.

It is worth noting that our method can be distinguished from other typical methods at the following aspects. First, the dilated convolution based model proposed by Yu and Koltun in \cite{yu2015multi} was designed for dense prediction problem without residual learning, which, however, is a powerful tool in our work. Second, Yu and Koltun recently presented a dilated residual network in \cite{yu2017dilated} for image classification and reported state-of-the-art performance. The main purpose of utilizing dilated convolution was to replace the pooling operations (2-stride convolution) in the original resNet \cite{he2016deep} to increase the resolution of the network\textquotesingle s output. While our work is inspired by \cite{yu2017dilated}, we focus on image denoising which is essentially different as classification problem. The main purpose of adopting dilated convolution in our work is to enlarge receptive field. Third, Zhang, et al. creatively proposed a residual denoiser (DnCNN) in \cite{zhang2017beyond}, and achieved state-of-the-art results for image denoising. However, dilated convolution was not considered, and the receptive field expansion was limited. In contrast, our model can obtain a more aggressive expansion of receptive field as compared in section \ref{DenoisingResNet}. While our method does not outperform DnCNN on denoising effect, our model has fewer layers and the computational cost can be effectively reduced. Although Zhang et al. also incorporated dilated convolution in their recent work \cite{zhang2017learning} for image denoising, they chose inconsistent dilation factors without considering the impact of extra zero paddings. As shown in section \ref{experiment}, our method outperforms their model.

The main contributions of this work are generalized in two-folds. Firstly, we propose a dilated residual network for Gaussian image denoising, which is computational efficient and outperforms most of the state-of-the-art models. Secondly, we introduce an approach to calculate receptive field size when dilated convolution is included. The rest of the paper is organized as follows. In section \ref{relatedwork}, we review the related literature. We present the dilated residual denoiser in section \ref{Method}. Extensive experiments and evaluation results can be found in section \ref{experiment}. Finally, we conclude our work in section \ref{conclusion}.

\section{Related Work}
\label{relatedwork}

In this section, we review the related literature regarding to dilated convolution, deep residual learning, and existing deep learning methods for image denoising.

\subsection{Deep Learning for Image Denoising}

Burger et al. \cite{burger2012image} concluded that multi-layer perceptron (MLP) can achieve superior image denoising performance with proper network depth, patch size, and training set. Xie et al.~\cite{xie2012image} proposed to combine sparse coding and pre-trained deep neural networks. Jain et al.~\cite{jain2009natural} demonstrated that the convolutional networks can provide comparable or better performance compared with the Markov random field (MRF) methods. Chen et al.~\cite{chen2017trainable} presented a trainable nonlinear reaction diffusion (TNRD) model. Specifically, the filters and the influence functions can be learned together from the training data. Lefkimmiatis et al.~\cite{lefkimmiatis2016non} coupled the convolutional network training with the non-local self-similarity property modeling with clean mapping as the output. Vemulapalli et al. \cite{vemulapalli2016deep} proposed a model which can handle a variety of noise levels by explicitly modeling the input noise variance. Zhang et al. \cite{zhang2017beyond} explored the feasibility of connecting residual learning and image denoising, and they trained CNN based denoisers as priors in \cite{zhang2017learning}. These methods reported more competitive performance than the traditional prior modeling based methods. And we move forward one more step by considering both denoising effect and computational cost.

\subsection{Dilated Convolution}

Dilated convolution was originally applied for wavelet decomposition \cite{holschneider1989real, shensa1992discrete} in signal processing in 1980\textquotesingle s. It supports exponential expanding of receptive field. Yu and Koltun developed a convolutional network for dense prediction in \cite{yu2015multi}, in which dilated convolution was adopted to systematically combine multi-scale contextual information without sacrificing resolution or coverage. It improved the accuracy of the state-of-the-art semantic segmentation method. The success can be mainly attributed to the expansion of receptive field by dilated convolution, and more relevant information can be perceived by CNN. Their work provides a simple yet effective way to enlarge receptive field for CNN. 

\subsection{Deep Residual Network}
Deep residual learning was initially proposed by He et al. in \cite{he2016deep} and it became the state-of-the-art method for image classification. The reliability has also been validated in medical image processing by Xu et al. in \cite{xu2017deep}. Zhang et al.~\cite{zhang2017beyond} discovered that the modeling of noisy image can share similar interpretation as the residual learning problem, they hence developed a residual model to learn noisy mapping for image denoising task. While their method brings state-of-the-art effect, our work will reveal that there is still leeway to reduce the computational burden and maintain a comparable denoising effect, when dilated convolution is combined. Yu and Koltun in \cite{yu2017dilated} extended residual learning by incorporating dilated convolution for image classification. Inspired by their work, we propose a dilated residual network for image denoising problem.

\section{The Proposed Method}
\label{Method}

\begin{figure*}
    \centering
    \includegraphics[width=0.9\textwidth]{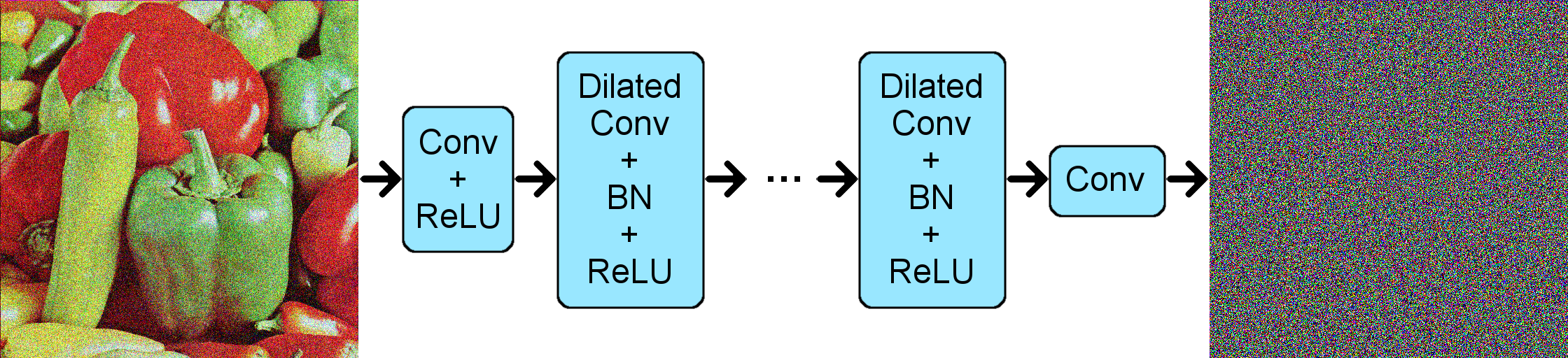}
    \caption{The network architecture with dilated convolution.}
    \label{figure1}
\end{figure*}

\begin{table*}
\centering
\caption{Receptive field size of our model and DnCNN \cite{zhang2017beyond}. }
\label{tablei}
\renewcommand\arraystretch{1.8}
\begin{tabular}{|l|ccccccccccccccccc|} 
\hline
\diagbox{Method}{Layer} & 1 & 2 & 3 & 4 & 5 & 6 & 7 & 8 & 9 & 10 & 11 & 12 & 13 & 14 & 15 & 16 & 17\\ 
\hline

our model              & 3  & 7  & 11  & 15  & 19  & 23  & 27  & 31  & 35  & (37)  & -  & (45)  & -  & -  & -  & -  & -  \\

DnCNN                  & 3  & 5  & 7  & 9  & 11  & 13  & 15  & 17  &  19  & 21  & 23  & 25  & 27  & 29  & 31  & 33  & 35 \\
\hline
\end{tabular}
\end{table*}

 Our dilated residual denoiser is capable of more effectively expanding receptive field and attaining a promising result of image denoising. Before moving forward to the proposed network architecture, we would like to discuss the impact of receptive field in CNN and why dilated convolution is suitable. By revisiting the relevant convolution arithmetic, we then introduce an approach to calculate receptive field size when dilated convolution is involved. In the end, our dilated residual denoiser will be described and we will also compare the proposed network with the original residual denoiser \cite{zhang2017beyond} in terms of receptive field.

\subsection{Receptive Field}

In convolutional nets, a pixel value in the output only depends on a certain region of the input. This region is known as the receptive field. Intuitively, a larger region of the input can capture more context information. Therefore, larger receptive field is desired for CNN so that no important image features are ignored. In general, there are two ways to increase receptive field. The most straightforward option is to stack several convolutional layers, and receptive field can be expanded linearly as mentioned in \cite{luo2016understanding}. However, the expansion rate is related to the dimension of convolutional filters. It has been shown that $3 \times $3 filter is most effective and frequently used in CNN \cite{li2016filter}. Hence receptive field size is only enlarged by a factor of 2 at each layer. The other option is utilizing pooling operation. Although it can multiplicatively expand receptive field, it is not directly applicable to image denoising since the output must have the same size as the input. One may argue that the original input size can be maintained by using up-convolution after pooling layers as in \cite{long2015fully}, however, we empirically discover that pooling followed by up-convolution cannot achieve acceptable resolution for denoising problem.

\subsection{Why Dilated Convolution}
\label{PenalizedSoftmaxLoss}

The advantage of dilated convolution is being capable of capturing more image clues \cite{yu2017dilated, yu2015multi} by expanding receptive field. In fact, using large size filter can also enlarge receptive field. One may want to replace multiple dilated convolutional layers by a single convolutional layer with large size filter. For instance, two $3\times$3 filters with 2-dilation can be replaced by one $9\times$9 filter with 1-dilation. However, it is not recommended in practice. This can be explained from three perspectives. Firstly, stacking convolutional layers of small size filters needs multiple activation layers, which will add more non-linearity to make the model discriminative. It is desired in deep learning. Secondly, the number of model parameters can be greatly decreased by using smaller size filters \cite{simonyan2014very}. Thirdly, $3\times$3 has been proven the most effective filter size for natural images according to the covariance analysis theory \cite{li2016filter}. Therefore, instead of using larger size filters, we adopt smaller size filters with dilated convolution to enlarge receptive field in this work.

\subsection{Receptive Field Calculation}
\label{ConvolutionArithmetic}

It is acknowledged that layer settings, which include filter size, padding, stride, and dilation, can affect receptive field size. To better clarify the principle, we briefly revisit convolutional arithmetic \cite{dumoulin2016guide} and introduce an approach to calculate receptive field size according to given settings. It is known that the output size $o_l$ of a convolutional layer $l\in\{1,2..,n\}$ can be computed as follows \cite{dumoulin2016guide},

\begin{equation}
    o_l=\floor*{\frac{i+2p-k}{s}}_l + 1,
\end{equation}
where $i$ is the input size. The convolutional filter size, zero-padding, and stride are denoted by $k$, $p$, and $s$, respectively. We extend Eqn. 1 by incorporating a dilation factor $d$, and the updated output size can be given by Eqn. 2.

\begin{equation}
    o_l=\floor*{\frac{i+2p-k-d}{s}}_l + 1.
\end{equation}

Let $r_l$, $s_l$, and $d_l$ be the receptive field size, the stride, and the dilation factor of a convolutional layer $l$, respectively. 
The filter size $k$ is a pre-determined value for all convolutional layers. Our empirical finding is that the receptive field size of layer $l$ can be computed by 

\begin{equation}
    r_l=r_{l-1}+(k-1)*d_l*\prod_{j=1}^{l-1} s_{j}.
\end{equation}
Note that $r_1$ is 1. In denoising problem, $s_j$ will always be 1 since there is no pooling layer ($stride >$ 1). Hence, Eqn.3 can be simplified to

\begin{equation}
\label{rfsizeEqn}
    r_l=r_{l-1}+(k-1)*d_l.
\end{equation}

Therefore, increasing the dilation factor will lead a larger receptive field. Our work is based on this fact. It seems that simply increasing $d_l$ will result in an unlimited receptive field, which is imagined as the ideal situation. However, increasing $d_l$ will change the output size as indicated in Eqn.2. To remain the output size unchanged, zero-padding $p$ must also be increased. Nevertheless, additional padding will degrade network performance. In fact, $d_l = 2$ can already provide an ideal expansion of receptive field for image denoising. This is also indicated by our experiments in section \ref{experiment}.

\begin{figure*}
  \centering
  \subfloat[Noisy Input]
  {\includegraphics[width=0.325\textwidth]{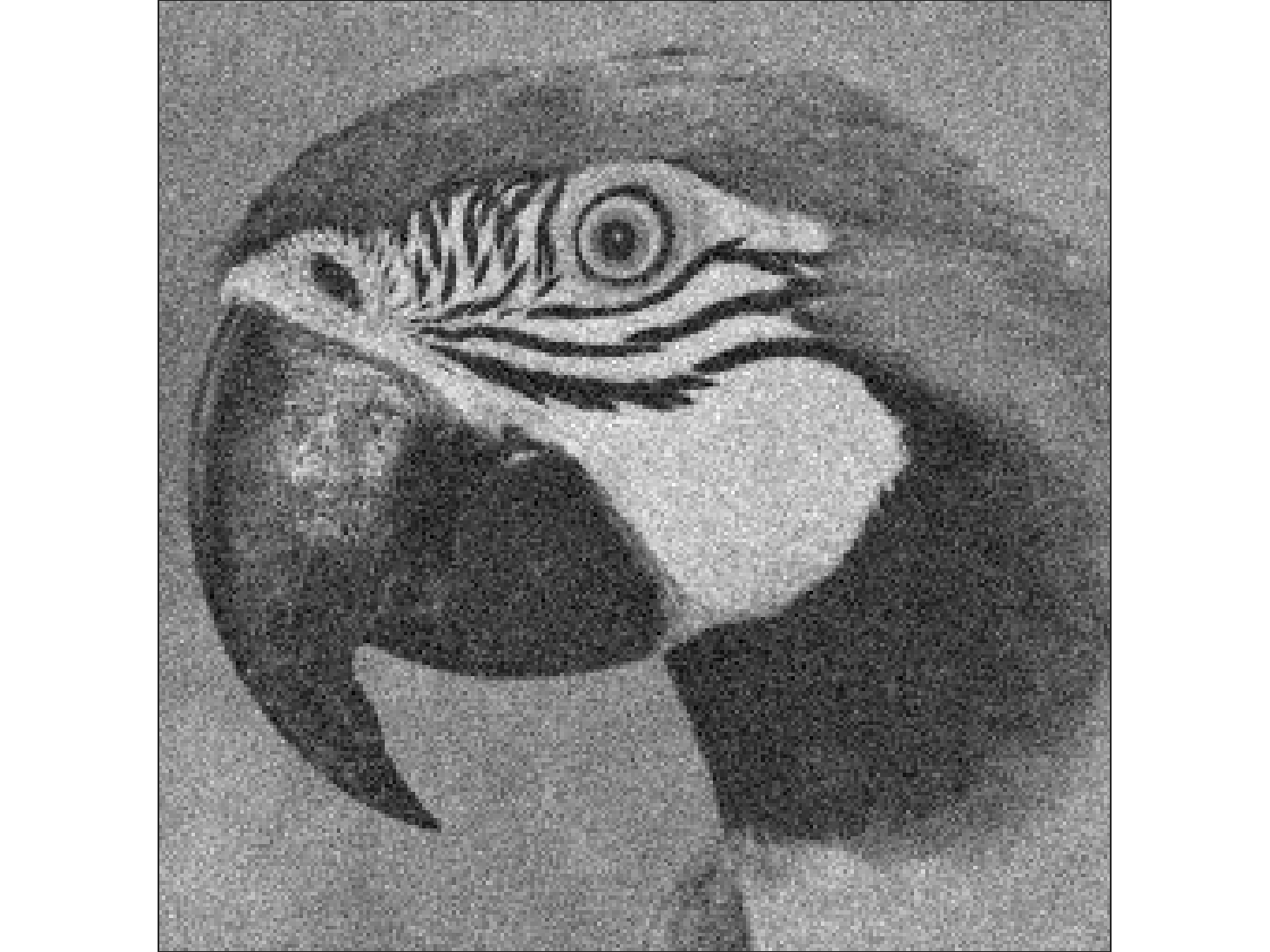}
  \label{map_a}}
  \hfill
  \subfloat[DnCNN]
  {\includegraphics[width=0.325\textwidth]{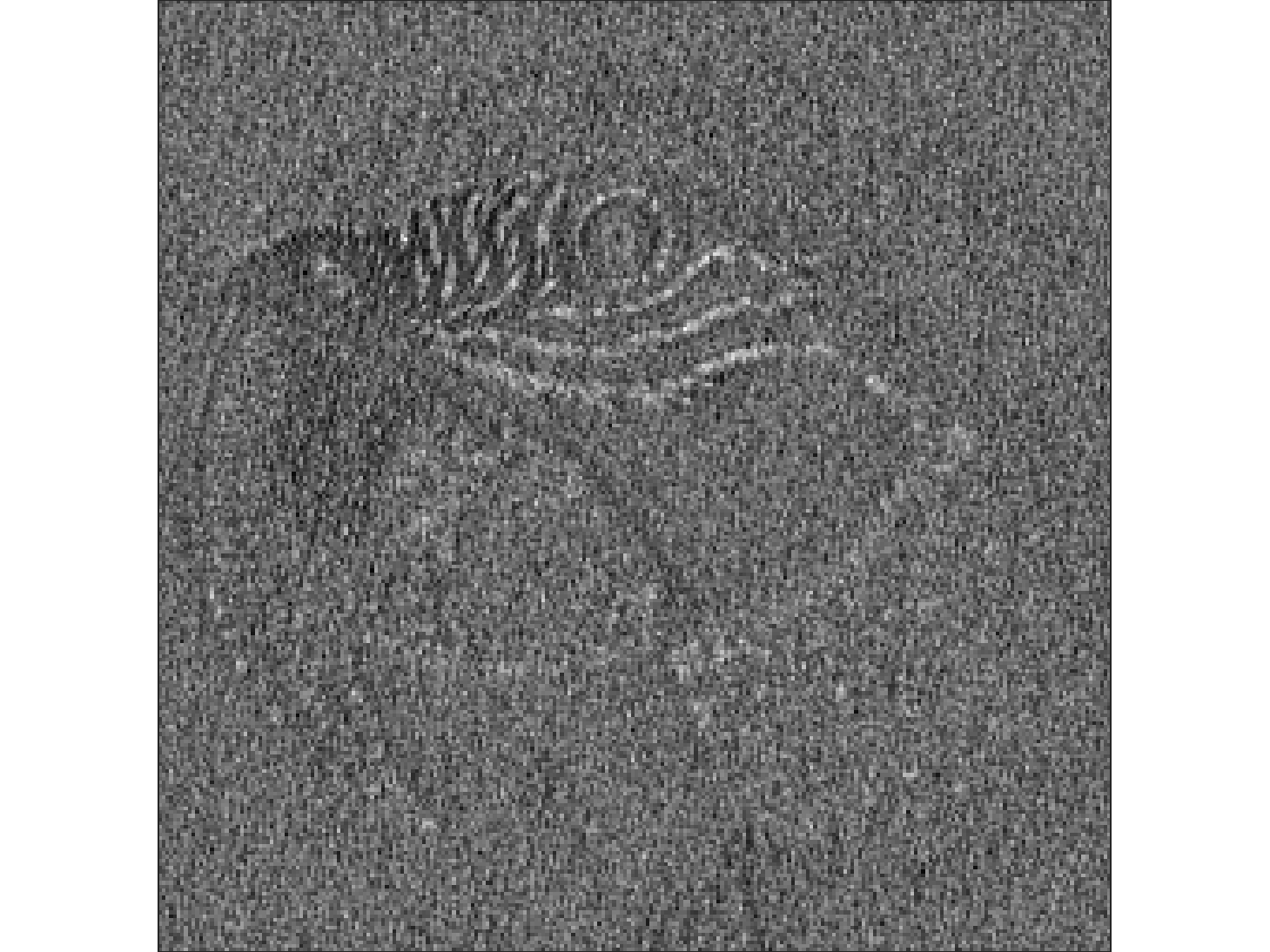}
  \label{map_b}}
  \subfloat[Ours]
  {\includegraphics[width=0.325\textwidth]{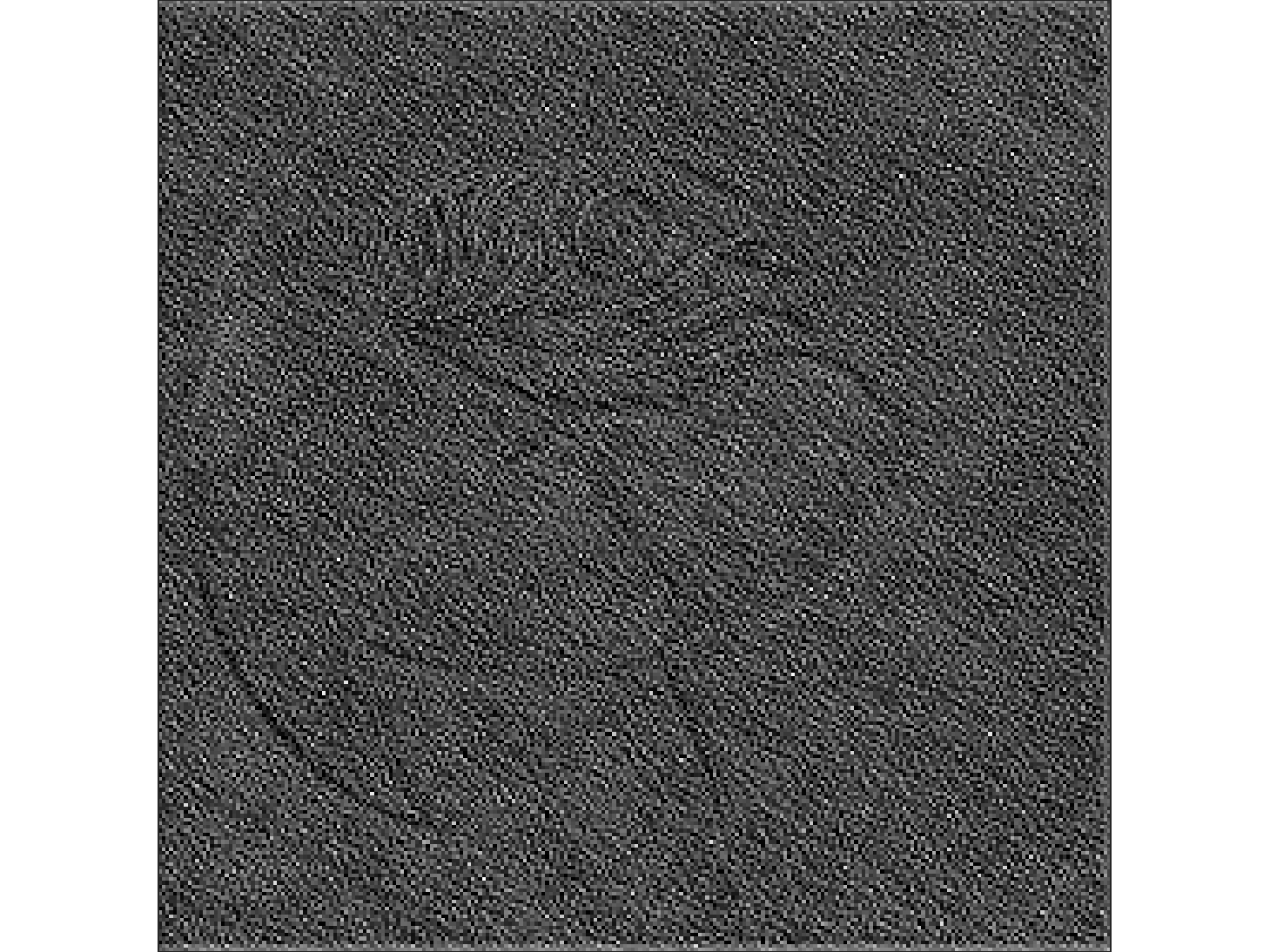}
  \label{map_c}}
  
  \subfloat[Noisy Input]
  {\includegraphics[width=0.325\textwidth]{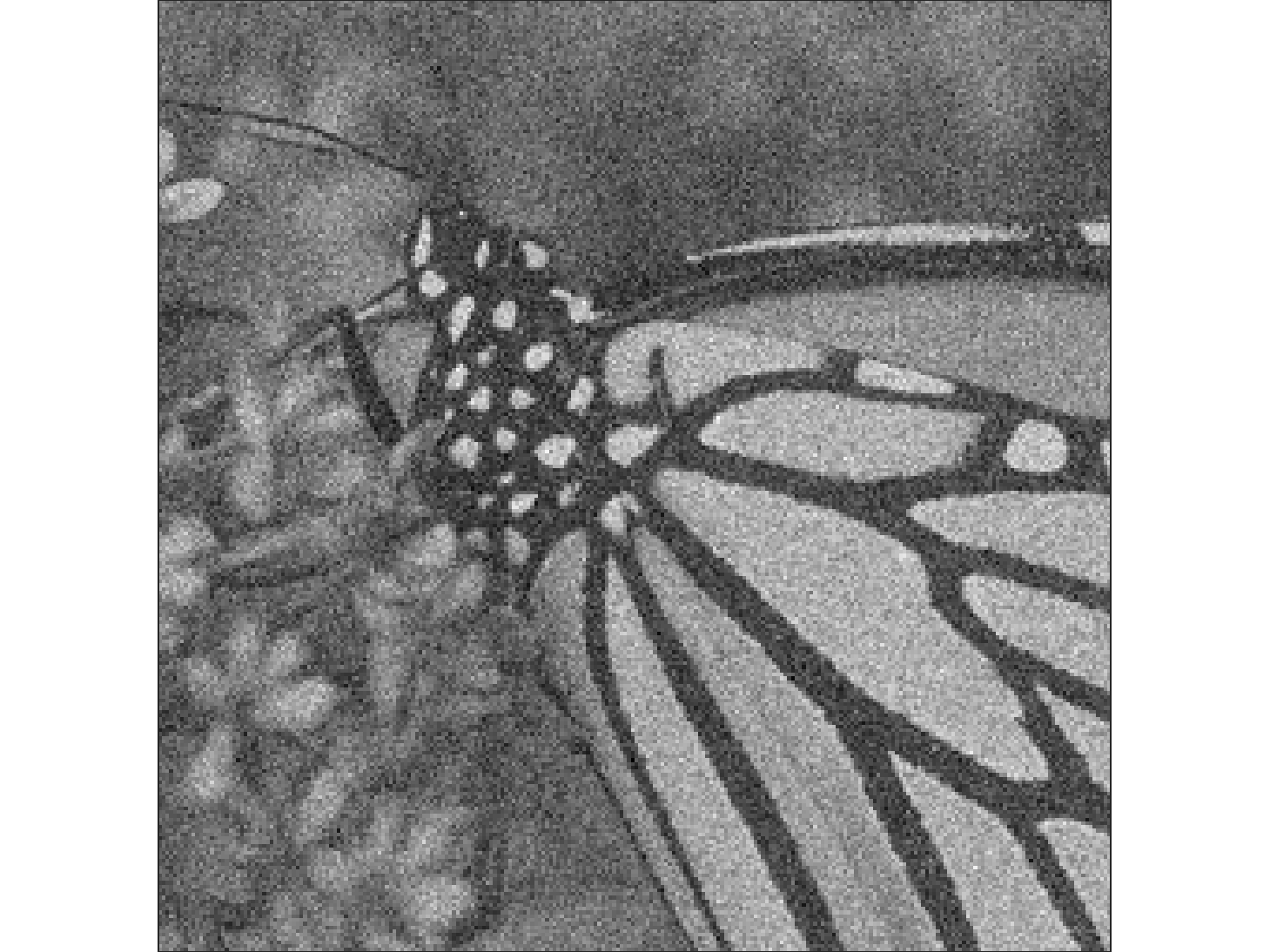}
  \label{map_e}}
  \subfloat[DnCNN]
  {\includegraphics[width=0.325\textwidth]{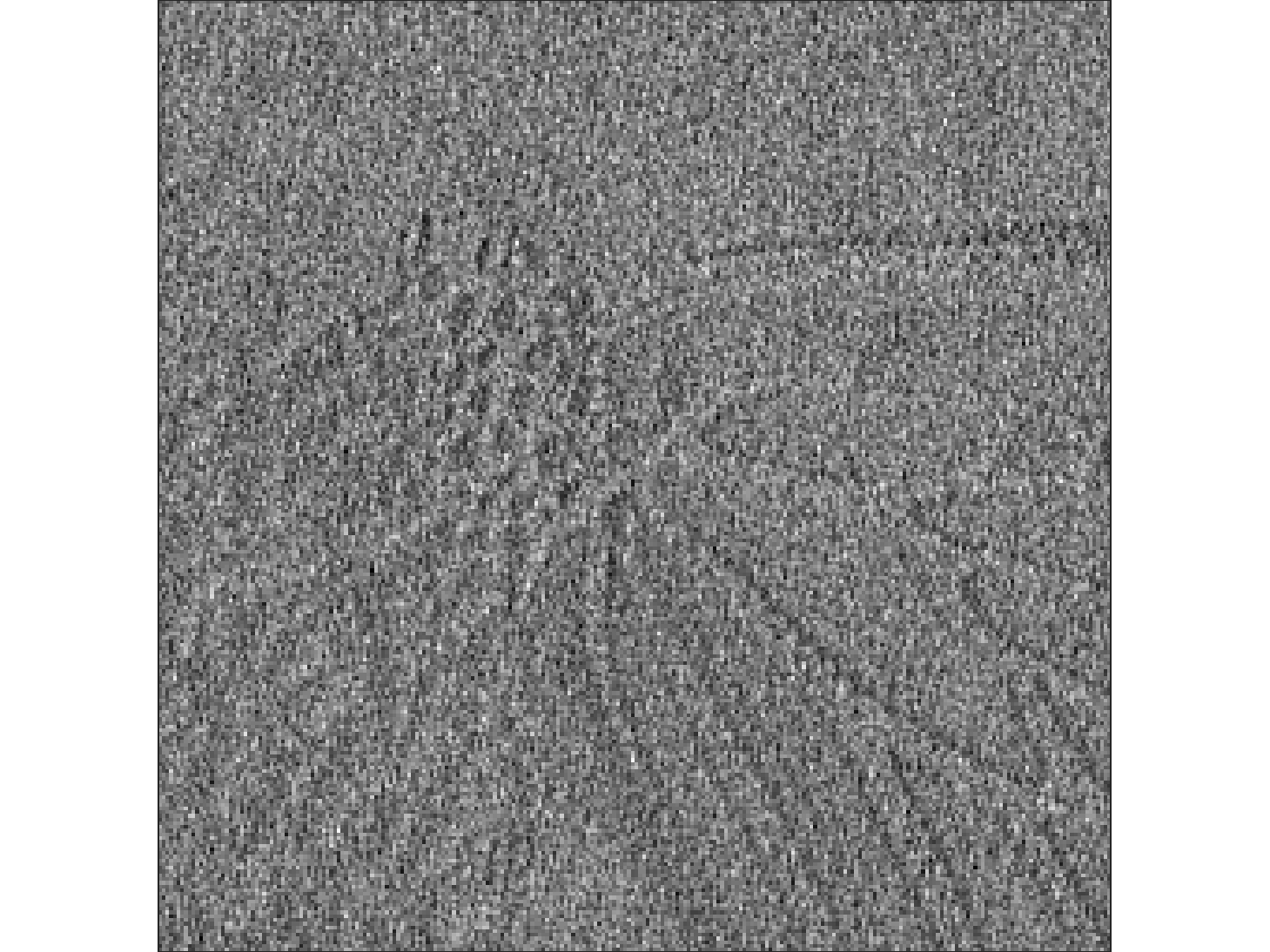}
  \label{map_f}}
  \subfloat[Ours]
  {\includegraphics[width=0.325\textwidth]{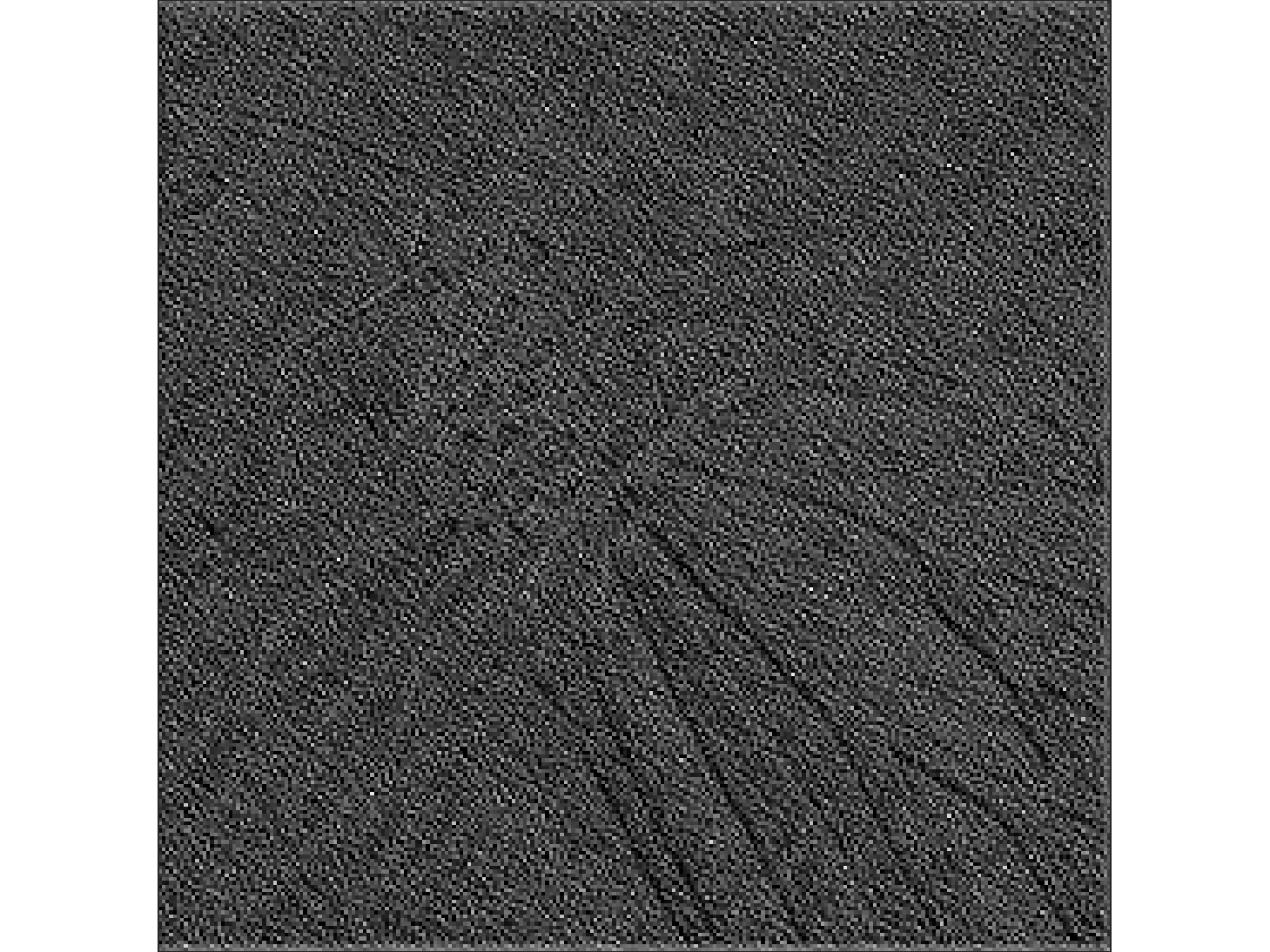}
  \label{map_g}}

  \caption{The highest average activation feature maps from the $10^{th}$ convolutional layer of DnCNN and our model.}
  \label{figureActivation}
\end{figure*}

\subsection{Dilated Residual Net for Denoising}
\label{DenoisingResNet}

Our model is based on residual learning aiming to learn a noisy mapping rather than a clean one. The architecture of our model is illustrated in Fig. \ref{figure1}, \textquoteleft DilatedConv-BN-ReLU\textquoteright ~is the fundamental block of our model. Batch normalization (BN) is used mainly for convergence purpose. On the other hand, it had been proven that BN and residual learning can benefit each other for image denoising \cite{zhang2017beyond}. Rectified linear unit (ReLU) is used to add non-linearity. Moreover, we employ dilated convolution to expand receptive field. The initial and the final convolutional layer still use regular convolution with dilation factor of 1, hence they are denoted by \textquoteleft Conv\textquoteright. \textquoteleft DilatedConv\textquoteright ~refers to the dilated convolution with factor of 2. To maintain the output size, zero-padding is set to 2 according to Eqn. 2. We choose 3 as the convolutional filter size as analyzed in section \ref{PenalizedSoftmaxLoss}. Moreover, classic CNN models \cite{he2016deep, szegedy2015going, simonyan2014very} all use $3 \times $3 filters, and Simonyan et al. in \cite{simonyan2014very} pointed out smaller filters impose implicit regularization on the network training. Since pooling operation is not included, the stride of each layer is set to 1. The L2 loss function shown in Eqn. 5 is used to learn the residual mapping, which measures the distance between the learned noisy mapping and the ground truth noise:

\begin{equation}
\label{equation4}
    L=\frac{1}{2N}\sum_{i=1}^N||f(y_i)-(y_i-x_i)||^2,
\end{equation}
where $f(y_i)$ is the learned noisy mapping of the latent clean image $x_i$ with respect to the noisy observation $y_i$, and $\{(y_i,x_i )\}_{i=1}^N$ represents a pair of noisy-clean image patch for training.

Furthermore, we compare the receptive field size of our model and DnCNN \cite{zhang2017beyond} in table \ref{tablei}. It shows that DnCNN increases receptive field size by a factor of 2 in each convolutional layer, whereas our model enlarges the size by a factor of 4 without impacting the output dimension. We use smaller image patch size $40\times$40 for gray image denoising and patch size $50\times$50 for color image denoising. Small patch size is shown to be more effective than larger patch by other works \cite{burger2012image, schmidt2014shrinkage, chen2017trainable, zhang2017beyond, zhang2017learning}. For gray image denoising, table \ref{tablei} shows that the $10^{th}$ convolutional layer of our model has a receptive field size of 37 (not 39, since the final layer uses 1-dilation), which is close to the patch size 40. Additional layers will make the receptive field size greater than or almost equal to the patch size, and the model may perceive irrelevant context which is not part of the image patch. Therefore, our network contains 10 convolutional layers for gray image denoising. Same principles can also be applied to determine the network depth for color image denoising. We take 12 convolutional layers for color image denoising and the receptive field size of the $12^{th}$ convolutional layer is 45 (patch size is 50). Our 10-layer model attains similar receptive field size as DnCNN \cite{zhang2017beyond} which adopted 17 convolutional layers for gray image denoising, and our 12-layer model has similar receptive field size as DnCNN which adopted 20 convolutional layers for color image denoising (receptive field size of the $20^{th}$ convolutional layer in DnCNN is 41, not shown in table \ref{tablei}). As a result, our network has fewer parameters and the computational burden can be reduced. The efficiency analysis will be given in table \ref{table3} and \ref{tableNEW}.

In addition, to validate the fact that dilated network can learn a desired mapping with lower depth, we give out the highest average activation feature maps of two arbitrary noisy images in Fig. \ref{figureActivation}. These maps are retrieved from the $10^{th}$ convolutional layer of DnCNN \cite{zhang2017beyond} and our model. The comparison illustrates that, with the same depth, our network can learn a residual mapping which contains more noise compared with DnCNN. Note that the $10^{th}$ convolutional layer is the final layer of our model for gray image denoising, whereas it is an intermediate layer of DnCNN. This indicates DnCNN needs more layers to compete with our results, which will bring more computational burden and model complexity. On the other hand, if DnCNN works with the same depth as our model, the denoising performance will be sacrificed. 
Since the receptive field size is different at the $10^{th}$ convolutional layer of DnCNN and our model, the trained filter parameters will be different. With the activation of ReLU, the amount of zero value pixels in our feature map differs from DnCNN. Therefore, the feature maps of our model and DnCNN show different gray levels. Dark background indicates more pixels are changed to zero value by ReLU.

\section{Experiments}
\label{experiment}

To validate the efficacy of our approach, we conduct experiments in Matlab using MatConvNet framework \cite{vedaldi2015matconvnet}. It has a convenient interface for designing network structure by adding or removing predefined layers. One Nvidia Geforce TITAN X GPU is used to accelerate the mini-batch processing. The testing time is measured on an Intel(R) Core(TM) i7-6700HQ CPU 2.60GHz. We train three networks for different tasks. The first one is for gray image denoising with specific noise levels. The second one is for color image denoising with specific noise levels, and the last one is for color image denoising with randomized noise levels (also known as blind denoising).

\subsection{Data Sets}

We follow the similar steps in \cite{zhang2017beyond} to collect the data. For gray image denoising with specific noise levels, we use 400 images of size $180\times$180 from Berkeley segmentation data set (BSD500) as training data and images are cropped to $128\times$1600 patches each of size $40\times$40. We apply three common noise levels, e.g., $\sigma =15, 25, 50$.  In addition, BSD68 data set, which consists of 68 gray images, is used for testing. No common images are shared between the training and the testing set. For color image denoising, 432 color images from CBSD500 are used for training, and the remaining 68 images are used for testing. There are total $128\times$3000 cropped patches, each of which has size $50\times$50. Three different noise levels $\sigma =15, 25, 50$ are applied to images. However, for blind denoising task, the noise levels are randomly selected from range [0,55].

\subsection{Compared Methods}

Besides comparing with the well-known prior modeling based methods such as BM3D \cite{dabov2007image}, LSSC \cite{mairal2009non}, EPLL \cite{zoran2011learning}, WNNM \cite{gu2014weighted}, we also consider other typical discriminative learning based approaches such as MLP \cite{burger2012image}, CSF \cite{schmidt2014shrinkage}, DGCRF \cite{vemulapalli2016deep}, NLNet \cite{lefkimmiatis2016non}, TNRD \cite{chen2017trainable}, DP \cite{zhang2017learning} and DnCNN \cite{zhang2017beyond}. All these methods reported very promising results, and DP and DnCNN reported leading results as they adopt residual learning which has been proven to be more suitable for image denoising.

\subsection{Network Training}
\label{training}

The initialization of filter weights is critical to network training. The most common option is random initialization. However, Yu and Koltun in \cite{yu2015multi} found that random initialization was not effective for their dilated network. Another popular method is usually called \textquoteleft Xavier\textquoteright ~initialization proposed by Glorot and Bengio in \cite{glorot2010understanding}. Nevertheless, in \cite{he2015delving}, He et al. pointed out the derivation of \textquoteleft Xavier\textquoteright ~initialization was based on linear activation, and it was not suitable for ReLU activation. Instead, they proposed a robust initialization by taking ReLU into consideration, also known as \textquoteleft MSRA\textquoteright ~initialization. It can make extreme deep model converge while \textquoteleft Xavier\textquoteright ~method cannot. In image classification, the state-of-the-art residual network \cite{he2016deep} was initialized by the \textquoteleft MSRA\textquoteright ~method. In addition, the residual denoiser proposed by Zhang et al. in \cite{zhang2017beyond} also adopted \textquoteleft MSRA\textquoteright ~initialization, and achieved very competitive results. As our model is essentially a dilated residual network, we adopt \textquoteleft MSRA\textquoteright ~method to initialize filter weights. 

The training is performed by stochastic gradient descent (SGD) algorithm with a momentum of 0.9 and a mini-batch size of 128. The initial learning rate is set to 0.001, and reduced to 0.0001 after 30 out of 40 epochs. Note that DnCNN trained 50 epochs, whereas 40 epochs are sufficient for our model to converge due to the fewer layers. The weight decay is set to 0.0001 to regularize the learned filters.

\begin{table}
\centering
\caption{The amount of parameters and the needed graphic memory size for loading the training data. The mini-batch size is 128. The color denoiser contains more parameters due to the deeper architecture. }
\vspace{0.05in}
\label{table3}
\begin{tabular}{|c|c|c|c|c|}
\hline
Methods  & gray/param & gray/mem & color/param & color/mem  \\ \hline

DnCNN & $5.6\times10^{5}$ & 2GB & $6.7\times10^{5}$  & 4GB  \\ \hline

Ours & $3\times10^{5}$ & 1GB & $3.8\times10^{5}$  & 2GB  \\ \hline

\end{tabular}
\end{table}

\begin{table}
\centering
\caption{The training and the testing time of DnCNN and our model. Noise ($\sigma =25$) is added to each training patch. An arbitrary test image is randomly selected from the testing set and the same noise is added.}
\vspace{0.05in}
\label{tableNEW}
\begin{tabular}{|c|c|c|c|c|}
\hline
Methods  & gray/train & gray/test & color/train & color/test  \\ \hline

DnCNN & 20.83hrs & 1.11s & 76.67hrs  & 2.46s  \\ \hline

Ours & 9.33hrs & 0.61s & 34.00hrs  & 1.42s  \\ \hline

\end{tabular}
\end{table}

\subsection{Results Analysis}

In our work, peak signal-to-noise ratio (PSNR) is adopted to evaluate denoising effect as it has been widely used to measure image quality. Moreover, it also avoids the problem of MSE (mean square error) which highly depends on image intensity scaling. For gray image denoising, we first compare our method with other well-known methods on BSD68 data set. The results are illustrated in table \ref{table2}. It can be seen that our model outperforms other methods and achieves comparable results with respect to the state-of-the-art DnCNN \cite{zhang2017beyond}, for all the three specific noise levels. While our method does not exceed DnCNN, our network contains fewer layers which require less computational cost. The amount of network parameters and the needed graphic memory size for loading the training data is given in table \ref{table3}. Note that the values are different for gray and color image denoising due to the different network depth. For instance, we use 10 convolutional layers for gray image denoising and 12 for color image denoising, whereas DnCNN takes 17 and 20, respectively. Moreover, the training and the testing time are illustrated in table \ref{tableNEW}. Both DnCNN and our models are trained with a specific noise ($\sigma =25$). For testing, we randomly select an image from the testing set, and add the same noise. DnCNN and our model are then applied to denoise the corrupted image. Note that we measure the training time by GPU and the testing time by CPU, since training is usually performed on GPU. However, on user's end, the device may not be equipped with independent GPU, therefore, measuring CPU-based testing time makes more sense.

\begin{table*}
\centering
\caption{The average PSNR of different methods on the gray version of BSD68 data set.}
\vspace{0.05in}
\label{table2}
\begin{tabular}{|c|c|c|c|c|c|c|c|c|c|c|c|c|}
\hline
\multicolumn{1}{|l|}{Methods} & \multicolumn{1}{l|}{BM3D} & \multicolumn{1}{l|}{MLP} & \multicolumn{1}{l|}{EPLL} & \multicolumn{1}{l|}{LSSC} & \multicolumn{1}{l|}{CSF} & \multicolumn{1}{l|}{WNNM} & \multicolumn{1}{l|}{DGCRF} & \multicolumn{1}{l|}{TNRD} & \multicolumn{1}{l|}{NLNet} &
\multicolumn{1}{l|}{DnCNN} & \multicolumn{1}{l|}{DP} &
\multicolumn{1}{l|}{Ours} \\ \hline
$\sigma$ = 15                        & 31.08                     & -                        & 31.21                     & 31.27                     & 31.24                    & 31.37                     & 31.43                      & 31.42                     & 31.52                 & 31.73          & 31.63              & 31.68            \\ \hline
$\sigma$ = 25                        & 28.57                     & 28.96                    & 28.68                     & 28.71                     & 28.74                    & 28.83                     & 28.89                      & 28.92                     & 29.03                      & 29.23            & 29.15            & 29.18          \\ \hline
$\sigma$ = 50                        & 25.62                     & 26.03                    & 25.67                     & 25.72                     & -                        & 25.87                     & -                          & 25.96                     & 26.07                      & 26.23           & 26.19             & 26.21            \\ \hline
\end{tabular}
\end{table*}

For color image denoising, we train our model with specific and randomized noise levels. Table \ref{table4} depicts the competency of our model trained with specific noise levels. Like gray image case, our method presents similar results to DnCNN, which is trained with specific noise levels as well. Nevertheless, our shallow network contains fewer parameters which require less time for training and testing. This has been verified in table \ref{tableNEW}. Moreover, our model also needs less graphics memory for loading the training data. The amount of parameters and the needed graphics memory for color image denoising are also given in table \ref{table3}. It indicates that our model is more economical without sacrificing denoising performance.

\begin{figure*}
  \centering
  \subfloat[Noisy/20.18dB]
  {\includegraphics[width=0.2432\textwidth]{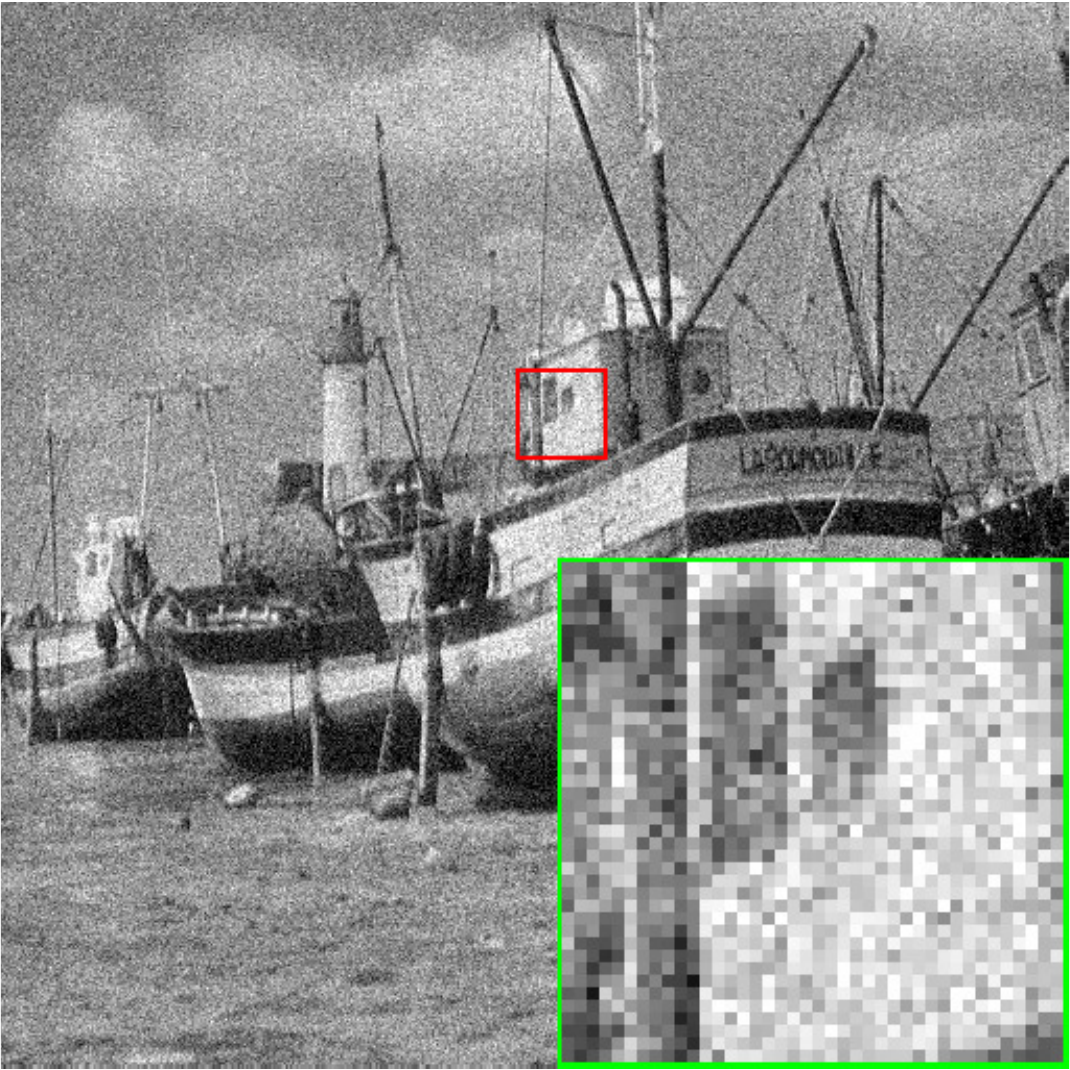}
  \label{fig2a}}
  \hfill
  \subfloat[BM3D/29.91dB]
  {\includegraphics[width=0.2432\textwidth]{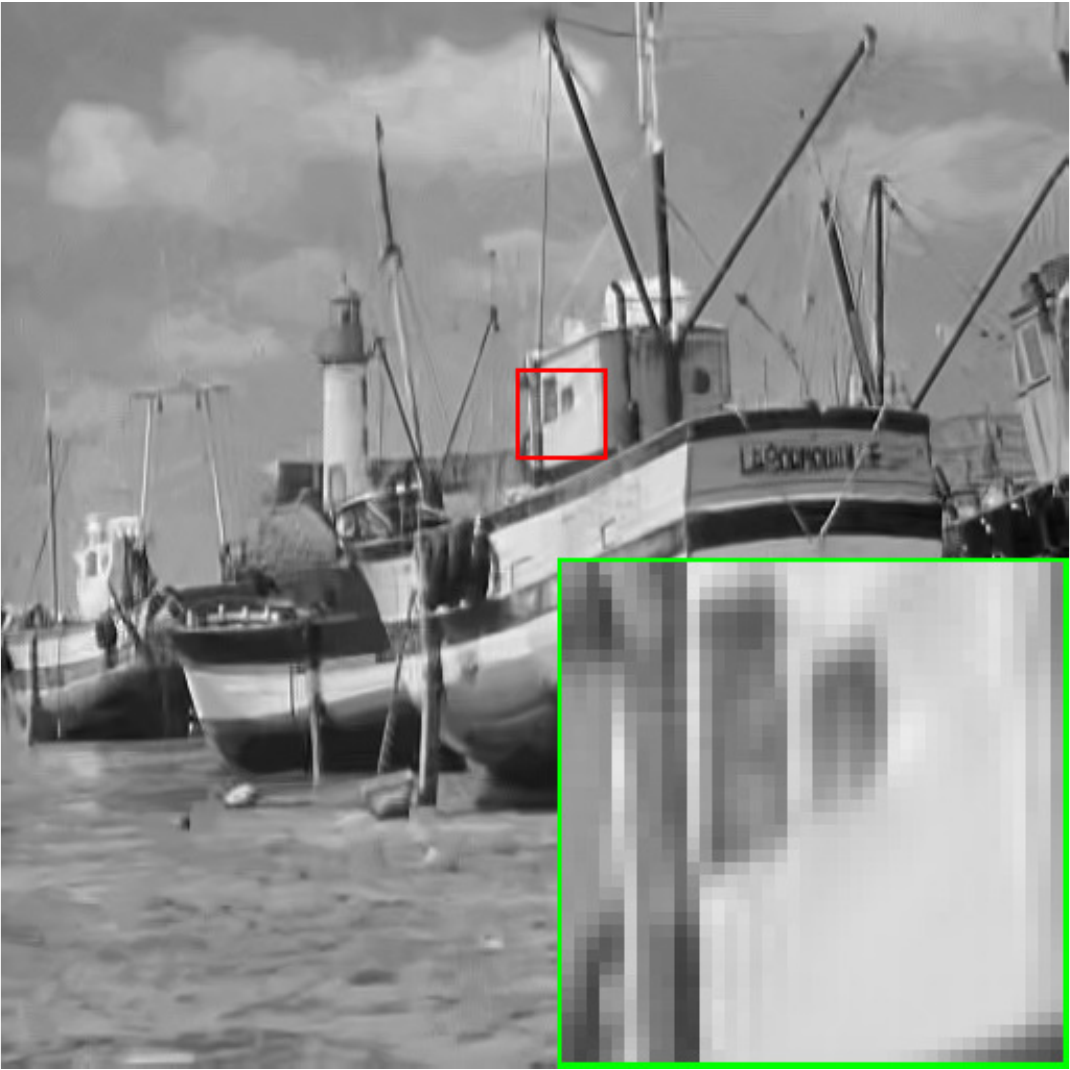}
  \label{fig2b}}
    \subfloat[EPLL/29.69dB]
  {\includegraphics[width=0.2432\textwidth]{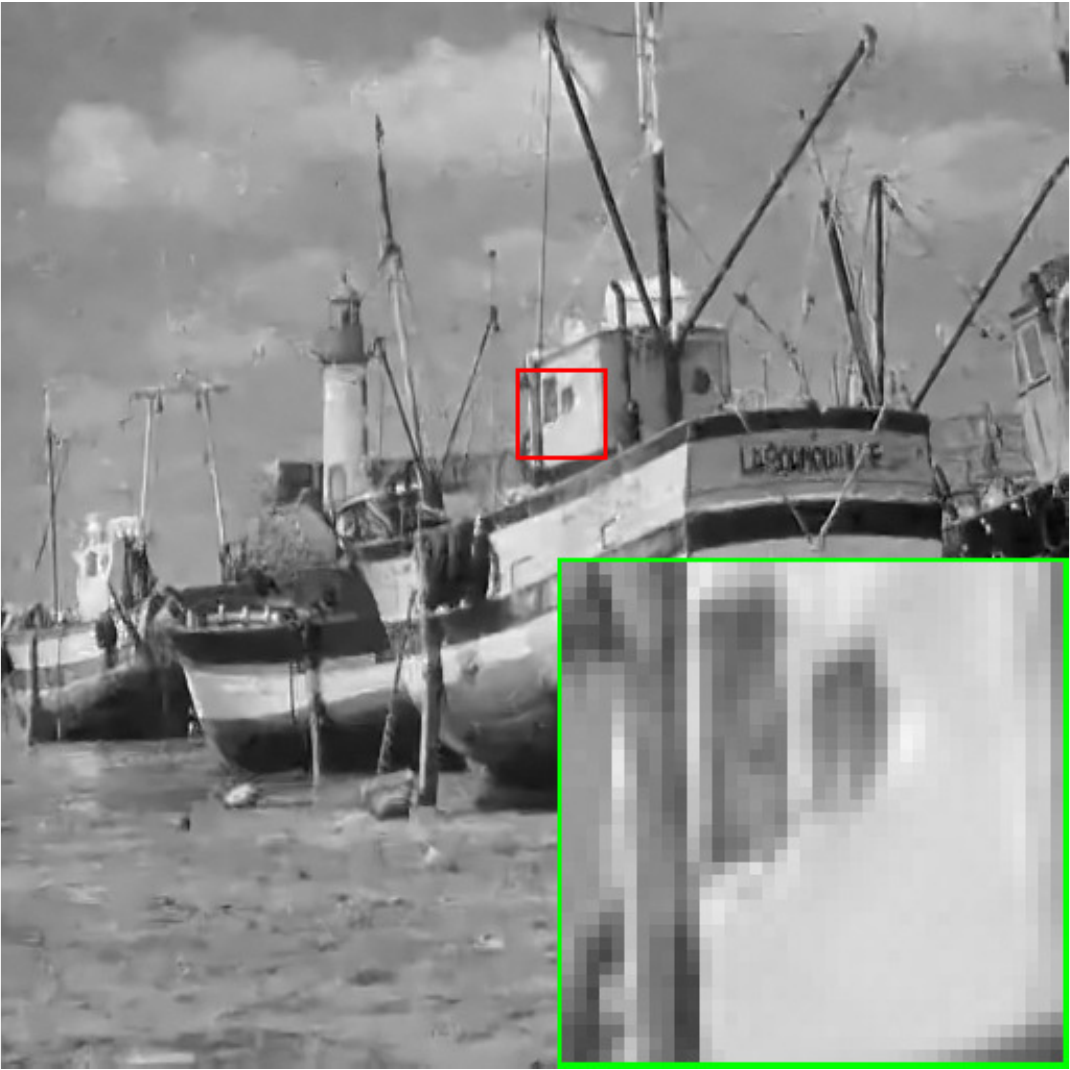}
  \label{fig2c}}
    \subfloat[MLP/29.95dB]
  {\includegraphics[width=0.2432\textwidth]{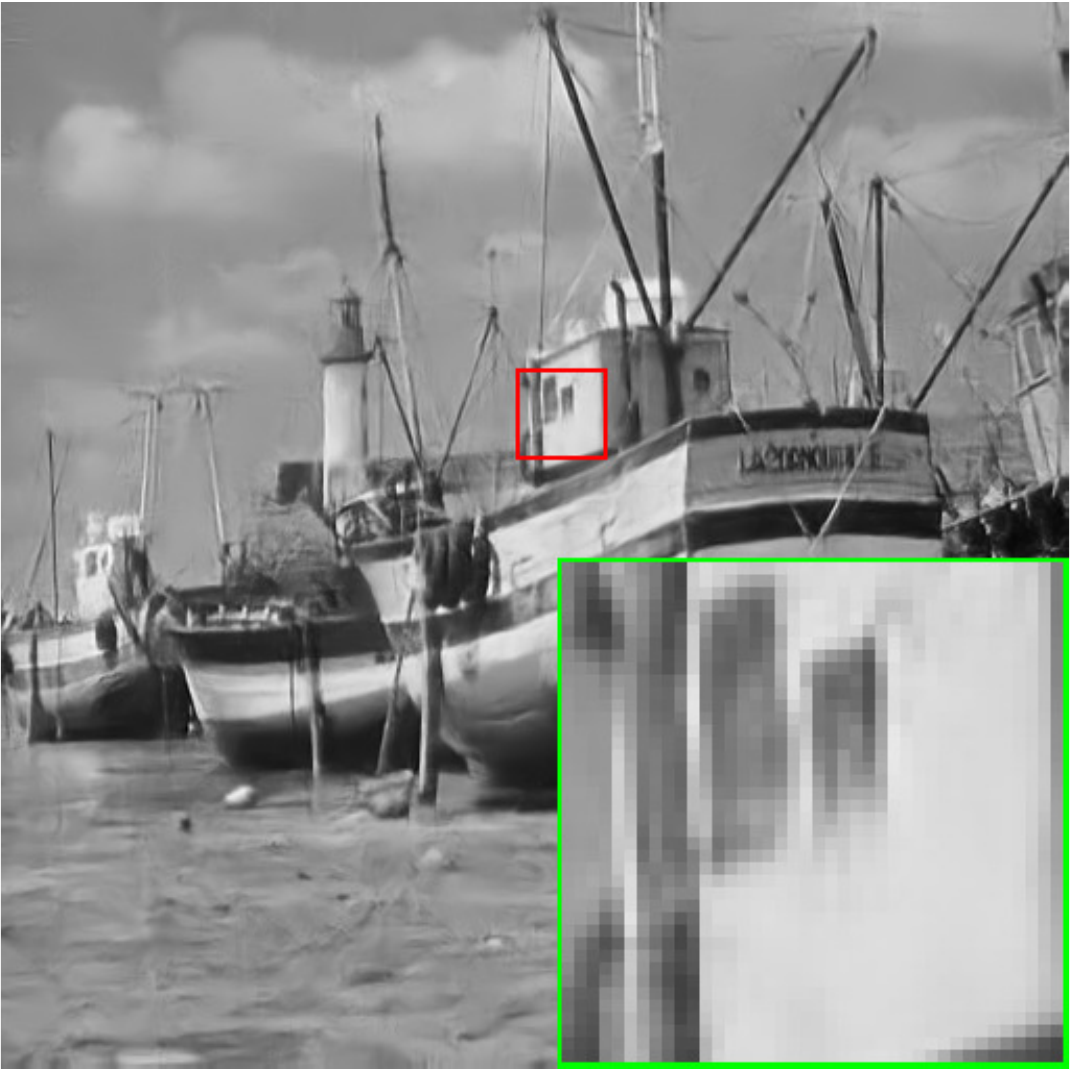}
  \label{fig2d}}
  
  \subfloat[TNRD/29.92dB]
  {\includegraphics[width=0.2432\textwidth]{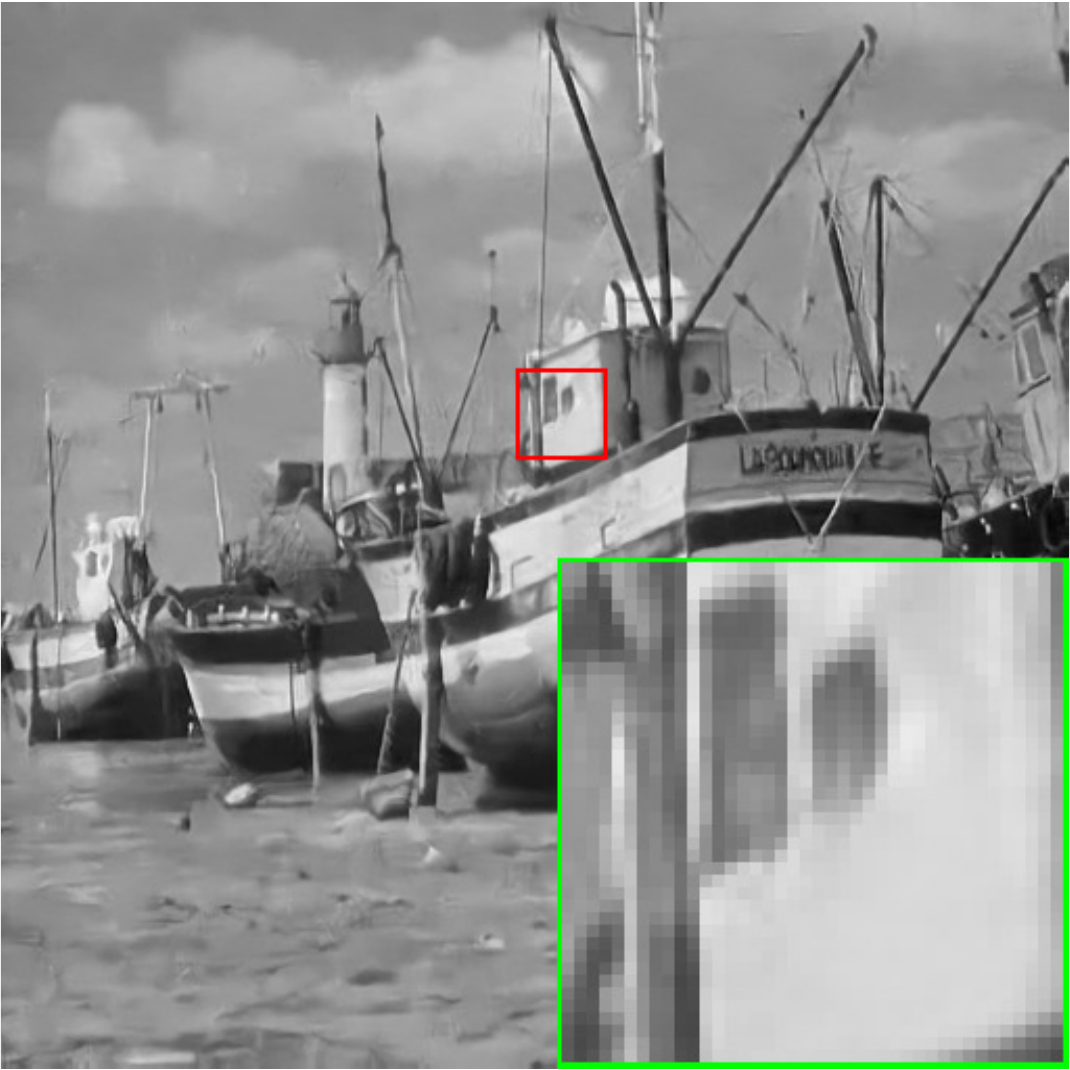}
  \label{fig2e}}
  \hfill
  \subfloat[WNNM/30.03dB]
  {\includegraphics[width=0.2432\textwidth]{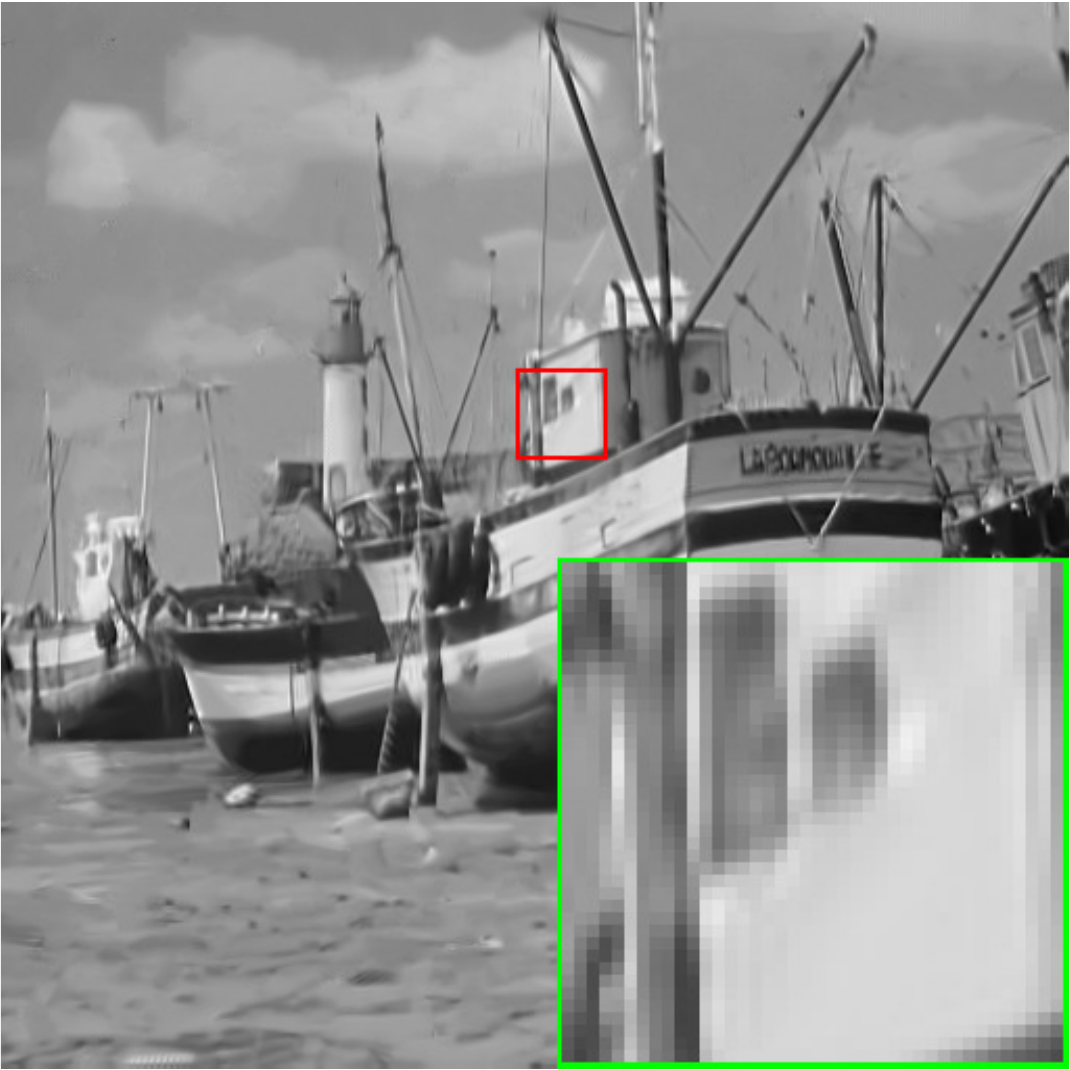}
  \label{fig2f}}
    \subfloat[DnCNN/30.22dB]
  {\includegraphics[width=0.2432\textwidth]{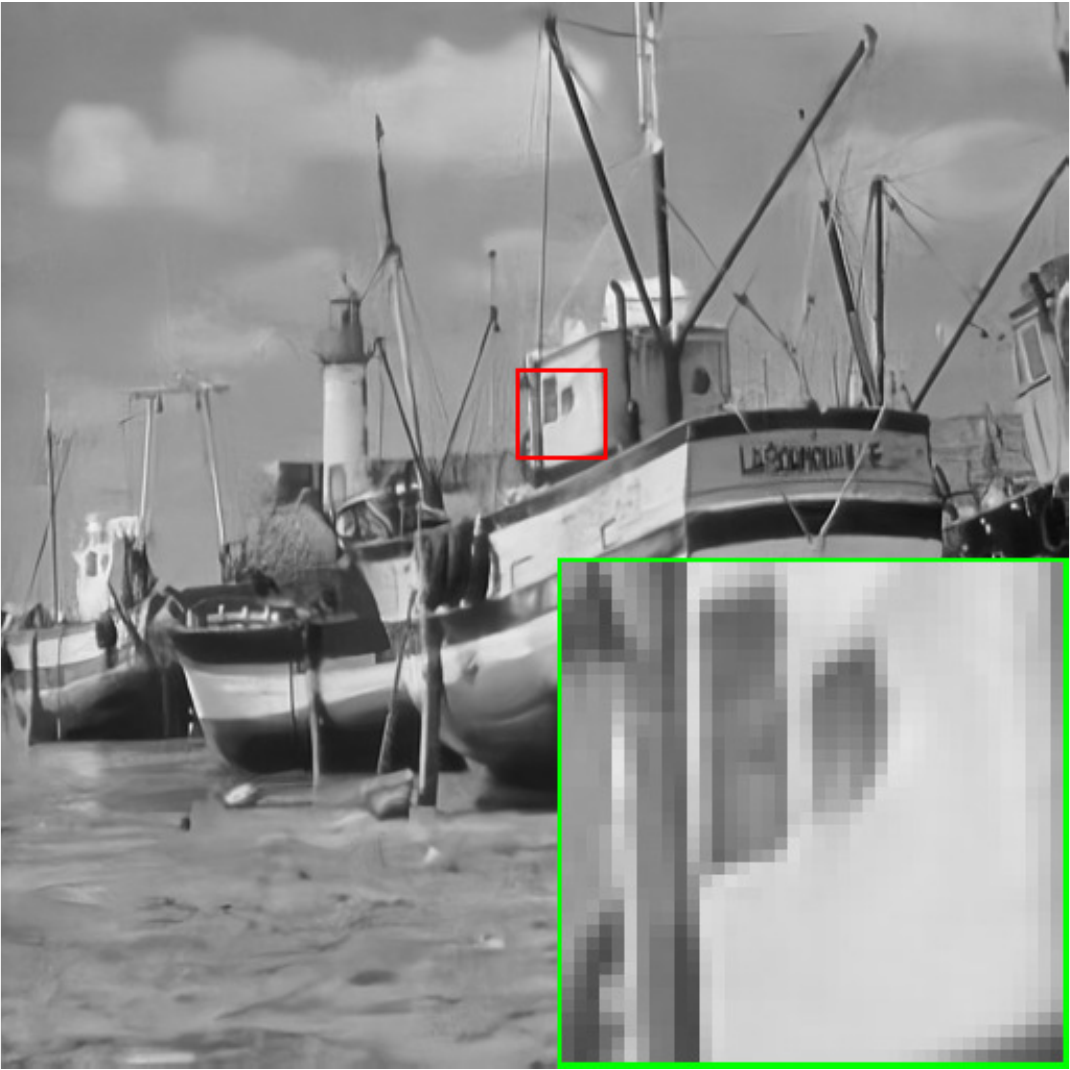}
  \label{fig2g}}
    \subfloat[Ours/30.17dB]
  {\includegraphics[width=0.2432\textwidth]{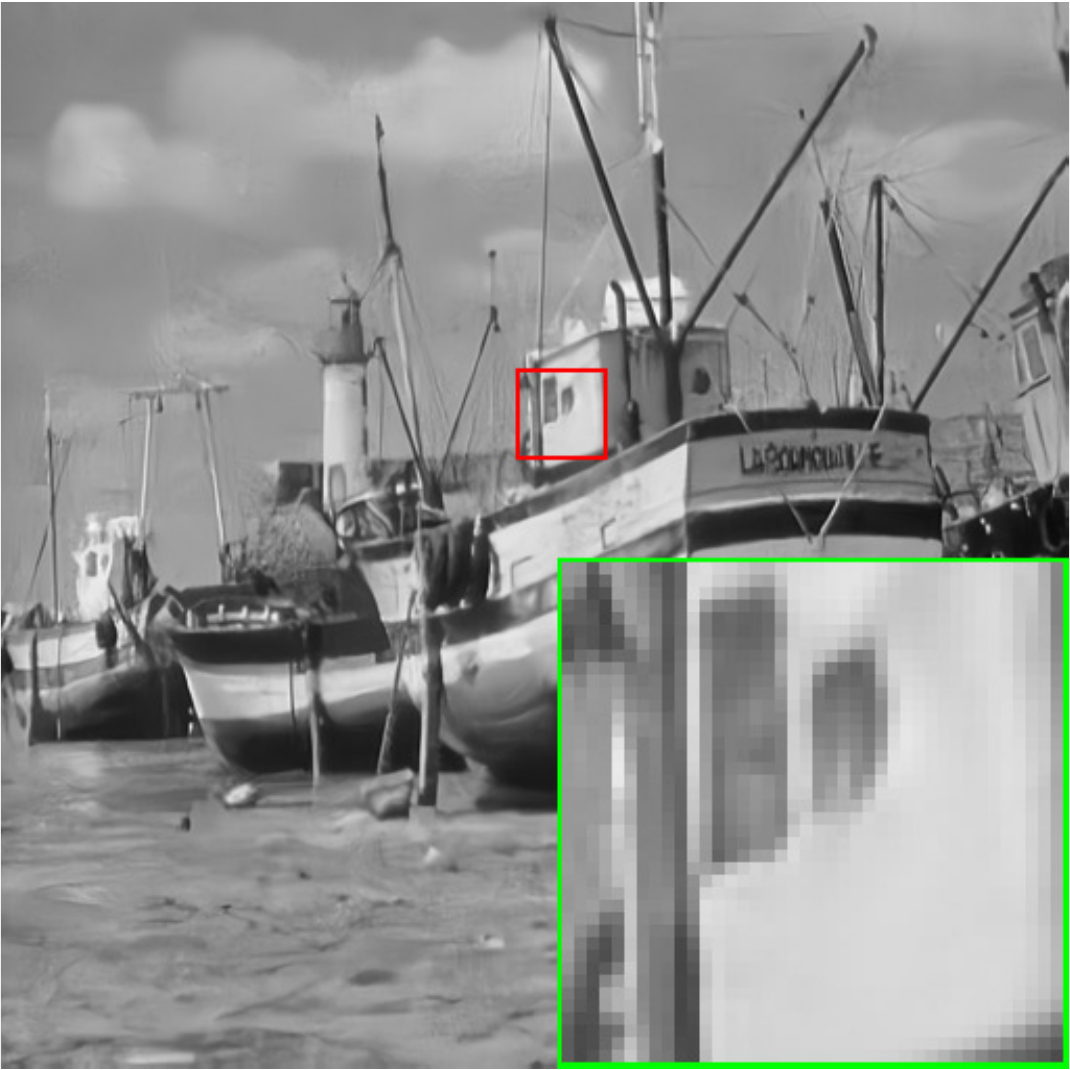}
  \label{fig2h}}

  \caption{Visual comparison of gray image denoising between our method and other popular methods. Our model is trained with a specific noise level ($\sigma=25$). The input image contains noise ($\sigma=25$).}
  \label{figure2}
\end{figure*}

The visual comparison between our method and other well-known methods is given in Fig. \ref{figure2} $\sim$ Fig. \ref{figure4}. We add noise ($\sigma =25$) to an arbitrary gray image, and our model is trained with the same noise level. The denoising effect of each method is shown in Fig. \ref{figure2}. Unfortunately, we are not able to give the visual denoising result for DGCRF \cite{vemulapalli2016deep} or NLNet \cite{lefkimmiatis2016non} since the codes are not released. It can be seen that our method gives more clear result than the other methods. Although it does not outperform DnCNN in terms of PSNR, the visual effect is very close to that of DnCNN. However, our model is more efficient than DnCNN as compared in table \ref{table3} and \ref{tableNEW}. 

\begin{figure*}
  \centering
  \subfloat[Noisy/17.70dB]
  {\includegraphics[width=0.2432\textwidth]{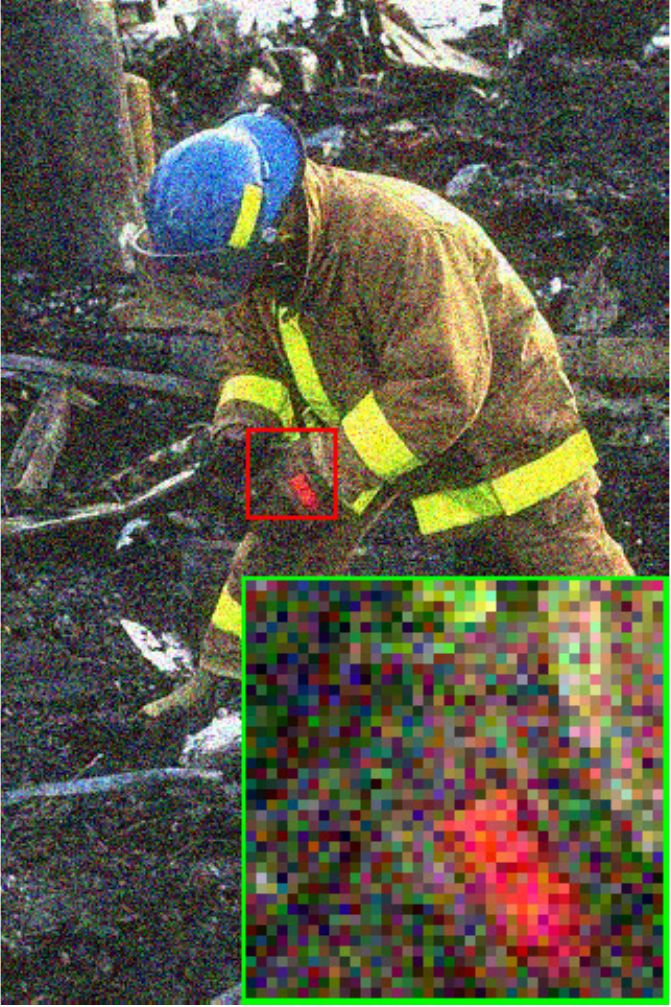}
  \label{fig3a}}
  \hfill
  \subfloat[CBM3D/27.23dB]
  {\includegraphics[width=0.2432\textwidth]{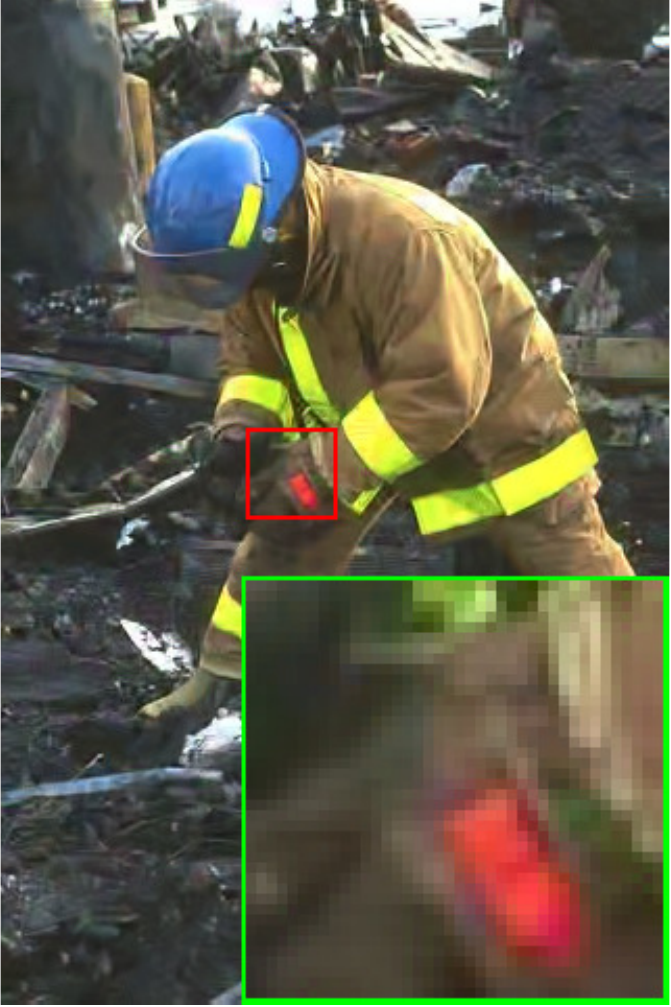}
  \label{fig3b}}
    \subfloat[DnCNN/28.18dB]
  {\includegraphics[width=0.2432\textwidth]{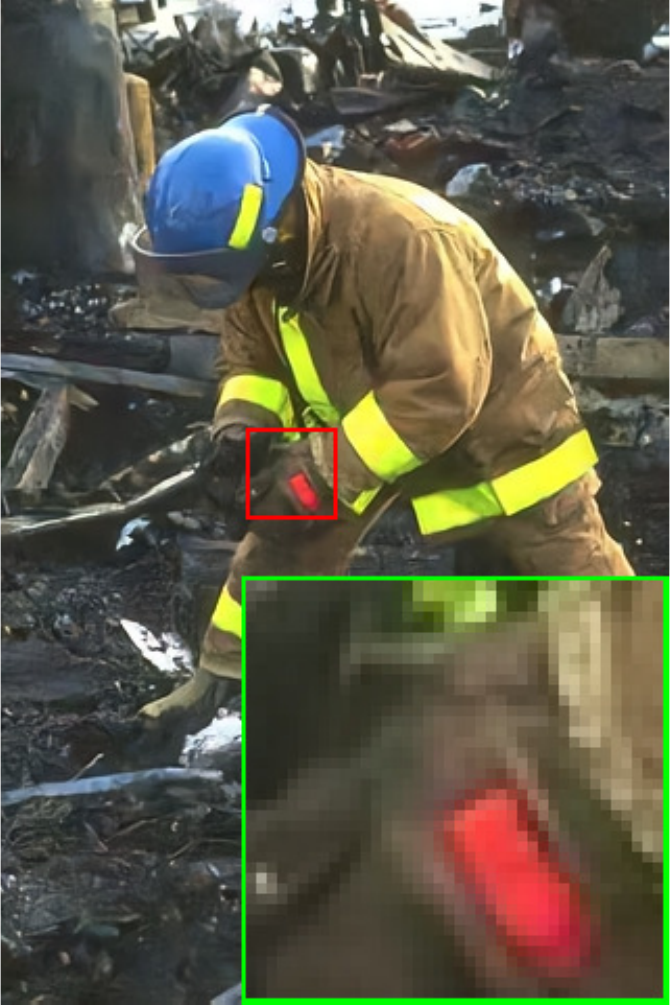}
  \label{fig3c}}
    \subfloat[Ours/28.14dB]
  {\includegraphics[width=0.2432\textwidth]{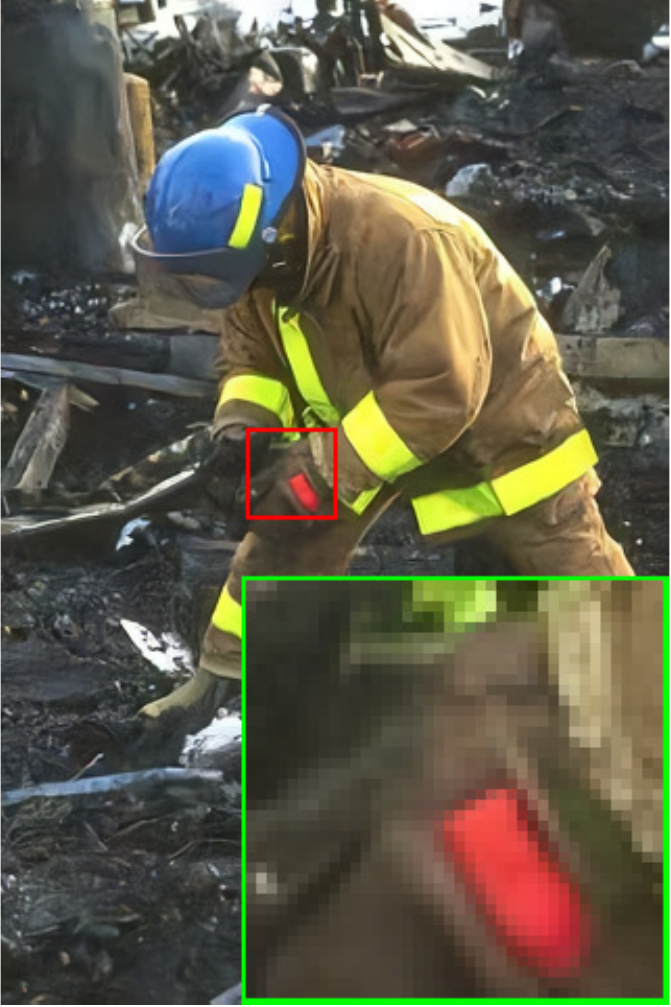}
  \label{fig3d}}
  
  \caption{Visual comparison of color image denoising between our method and CBM3D and DnCNN. Our model and DnCNN are trained with randomized noise levels from range [0,55]. The input image contains noise ($\sigma=35$).}
  \label{figure3}
\end{figure*}

\begin{figure*}
  \centering
  \subfloat[Noisy/15.10dB]
  {\includegraphics[width=0.2432\textwidth]{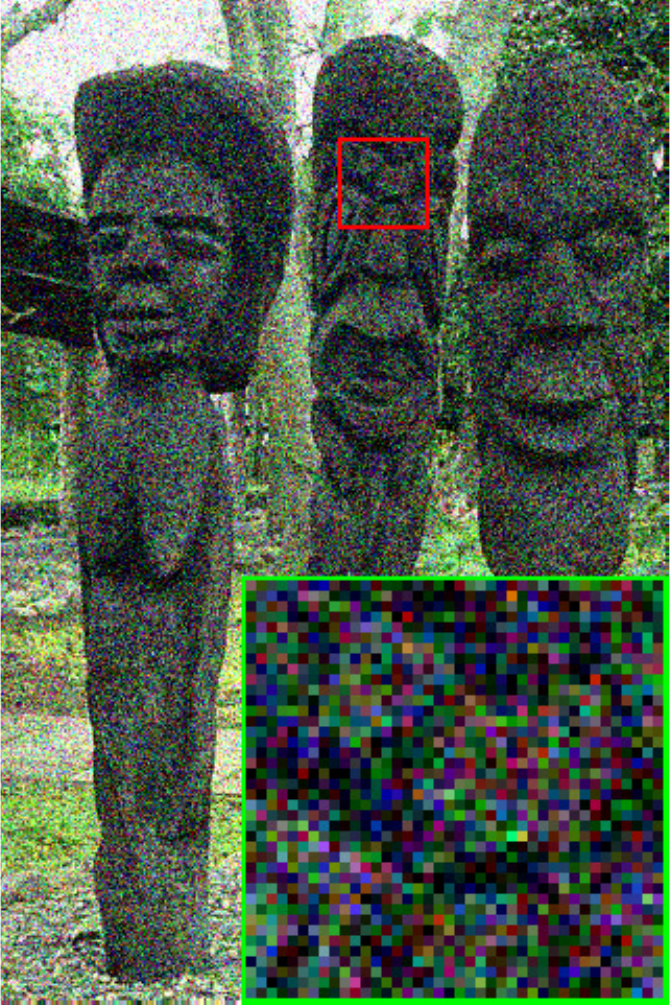}
  \label{fig4a}}
  \hfill
  \subfloat[CBM3D/24.32dB]
  {\includegraphics[width=0.2432\textwidth]{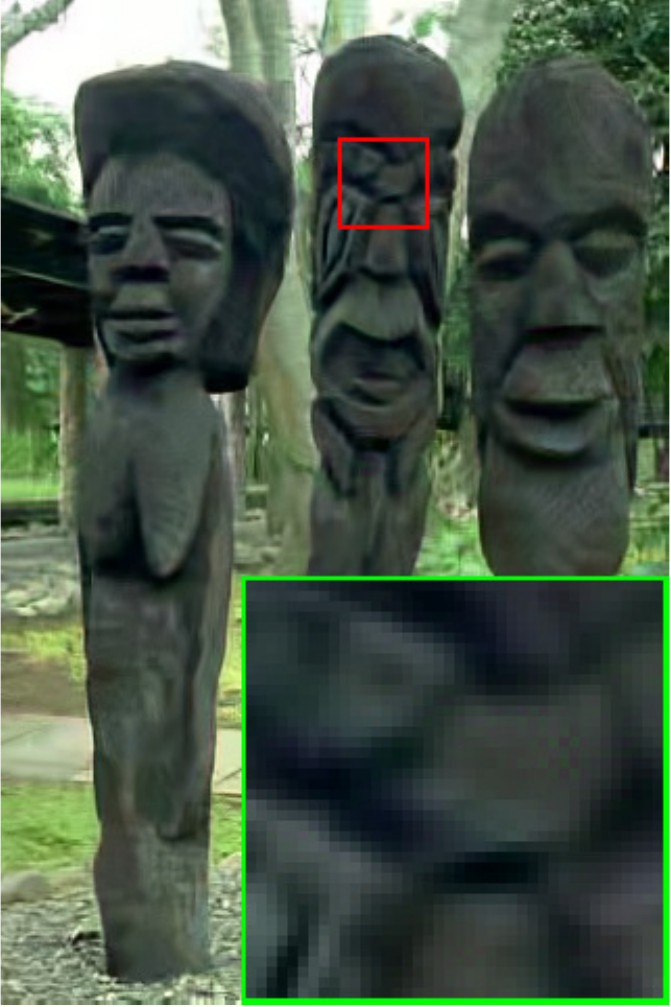}
  \label{fig4b}}
    \subfloat[DnCNN/24.97dB]
  {\includegraphics[width=0.2432\textwidth]{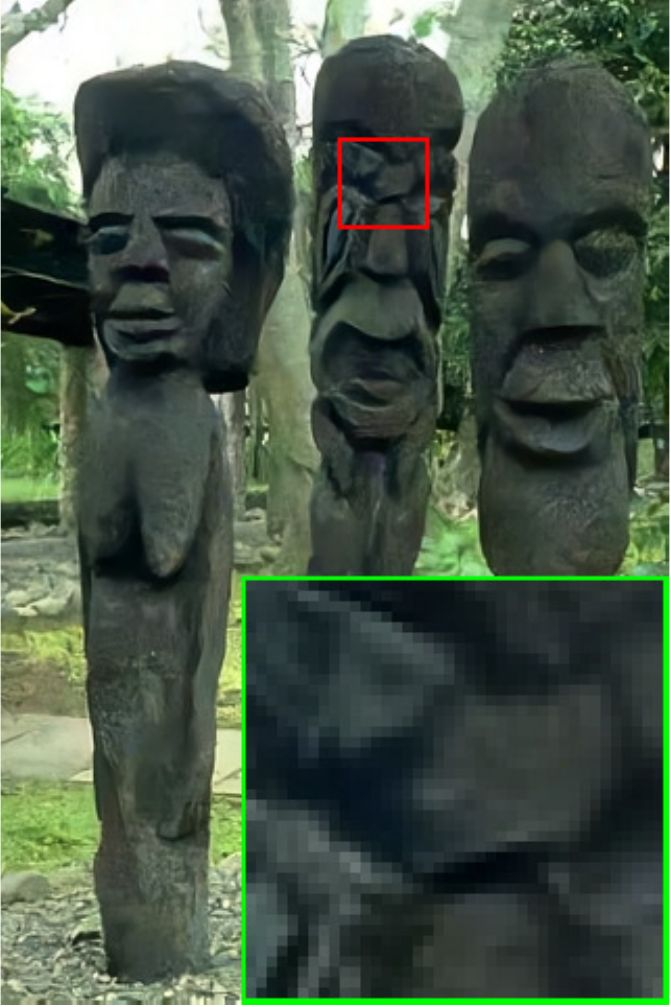}
  \label{fig4c}}
    \subfloat[Ours/24.92dB]
  {\includegraphics[width=0.2432\textwidth]{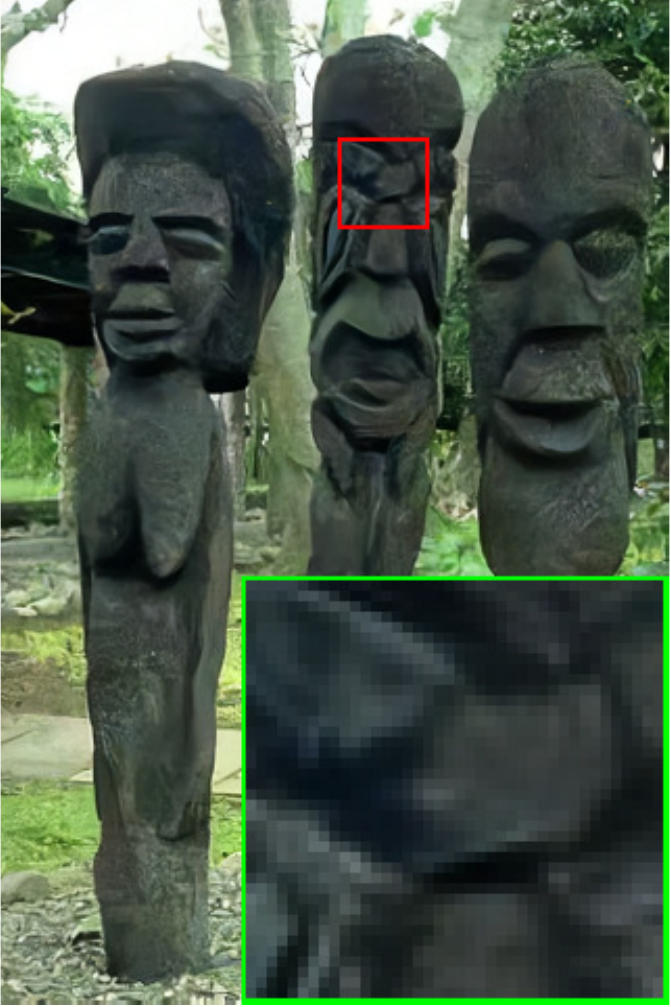}
  \label{fig4d}}
  
  \caption{Visual comparison of color image denoising between our method and CBM3D and DnCNN. Our model and DnCNN are trained with randomized noise levels from range [0,55]. The input image contains noise ($\sigma=50$).}
  \label{figure4}
\end{figure*}

While in Fig. \ref{figure3} and Fig. \ref{figure4}, to validate the effect of color image denoising with randomized noise levels, we add two different noise values ($\sigma = 35,50$) for each color image, respectively. Note that the visual comparison of color image denoising is only carried between our method and CBM3D \cite{dabov2007image} and DnCNN, since DnCNN, to our best knowledge, is the state-of-the-art method. Moreover, DnCNN also supports blind denoising. And CBM3D is the most widely used denoising approach in engineering. We compare our model with the version of DnCNN which is trained with randomized noise levels in the range of [0,55]. To have fair comparison, our model is also trained with randomized noise levels within the same range. Results indicate that our model restores image details much better than CBM3D, and the effect is also comparable to DnCNN.

\begin{table}
\centering
\caption{The average PSNR of different methods on the color version of BSD68 data set.}
\vspace{0.05in}
\label{table4}
\begin{tabular}{|c|c|c|c|c|c|c|}
\hline
Methods  & CBM3D & MLP   & TNRD  & DnCNN  & DP  &Ours   \\ \hline
$\sigma$ = 15 & 33.50 & - & 31.37  & 33.99 & 33.86  & 33.93 \\ \hline
$\sigma$ = 25 & 30.69 & 28.92 & 28.88  & 31.31 & 31.16  & 31.24 \\ \hline
$\sigma$ = 50 & 27.37 & 26.01 & 25.96  & 28.01 & 27.86  & 27.93 \\ \hline
\end{tabular}
\end{table}

\section{Conclusion}
\label{conclusion}

In this paper, we propose a dilated residual convolutional neural network for Gaussian image denoising. The dilated convolution is more effective to expand receptive field than simply stacking multiple layers. With fewer layers and less computational burden, our model is still comparable to the state-of-the-art denoising method. Moreover, we present an approach to compute receptive field size when dilated convolution is included. Extensive experiments show promising quantitative and visual results compared with other reputed denoising methods which are based on image prior modeling or discriminative learning.

  \bibliographystyle{unsrt}
  \bibliography{Reference} 

\end{document}